\documentclass[nonatbib,preprint,12pt,authoryear]{elsarticle}

\usepackage{amssymb}
\makeatletter
\let\c@author\relax
\makeatother
\usepackage{amsmath}
\usepackage{hyperref}
\usepackage{multirow}
\usepackage{tablefootnote}
\usepackage{graphicx}
\usepackage[table]{xcolor}
\usepackage{diagbox}
\usepackage{subcaption}
\usepackage{booktabs}
\usepackage{xcolor}
\usepackage[table]{xcolor}
\usepackage{bbm, dsfont}
\usepackage{apacite}
\usepackage{pifont}
\newcommand{\cmark}{\ding{51}}%
\newcommand{\xmark}{\ding{55}}%

\journal{Preprint}

\begin{document}
\newcommand{\specialcell}[2][c]{%
  \begin{tabular}[#1]{@{}c@{}}#2\end{tabular}}
\begin{frontmatter}



\title{Welcome New Doctor: Continual Learning with Expert Consultation and Autoregressive Inference for Whole Slide Image Analysis}


\author[aff1]{Doanh C. Bui}
\author[aff1]{Jin Tae Kwak}

\affiliation[aff1]{organization={School of Electrical Engineering, Korea University},
            city={Seoul},
            postcode={02841}, 
            country={Republic of Korea}}

\begin{abstract}
Whole Slide Image (WSI) analysis, with its ability to reveal detailed tissue structures in magnified views, plays a crucial role in cancer diagnosis and prognosis.
Due to their giga-sized nature, WSIs require substantial storage and computational resources for processing and training predictive models.
With the rapid increase in WSIs used in clinics and hospitals, there is a growing need for a continual learning system that can efficiently process and adapt existing models to new tasks without retraining or fine-tuning on previous tasks. Such a system must balance resource efficiency with high performance.
In this study, we introduce COSFormer, a Transformer-based continual learning framework tailored for multi-task WSI analysis. COSFormer is designed to learn sequentially from new tasks wile avoiding the need to revisit full historical datasets. We evaluate COSFormer on a sequence of seven WSI datasets covering seven organs and six WSI-related tasks under both class-incremental and task-incremental settings. The results demonstrate COSFormer’s superior generalizability and effectiveness compared to existing continual learning frameworks, establishing it as a robust solution for continual WSI analysis in clinical applications.
\end{abstract}




\end{frontmatter}


\section{Introduction}

Image classification is a long-standing problem in computer vision. Recent advancements in deep learning, particularly convolutional neural networks and Transformer-based methods, have significantly enhanced image classification performance across a wide range of applications at an unprecedented phase. 
In computational pathology, disease diagnosis, \textit{e.g.}, cancer grading or subtyping, can be formulated as an image classification problem \cite{noorbakhsh2020deep}, where computational methods analyze digitized tissue specimen images to determine disease status. However, image classification in computational pathology faces unique challenges: 1) \textit{Size}: it needs to handle gigapixel-sized whole slide images (WSIs), which are at least 10,000 times larger than standard input images in computer vision \cite{dimitriou2019deep}; 2) \textit{Variety}: A wide range of image classification tasks exist in pathology \cite{nguyen2024camp}. For instance, cancer diagnosis involves various objectives such as detecting cancer presence and metastasis, identifying histological grades \cite{nir2018automatic,le2021joint,bui2024spatial} or subtypes \cite{abmil,clam,dtfd,transmil}, and predicting treatment response \cite{zhang2020predicting}, recurrence \cite{pinckaers2022predicting,eminaga2024artificial}, or survival rates \cite{advmil,wetstein2022deep}; 3) \textit{Ambiguity}: While tissue characteristics are shared across different tasks and organ types, the designated class labels vary \cite{noorbakhsh2020deep}. Formulating each classification task as an independent multi-class problem neglects the common underlying knowledge of tissue structures; 4) \textit{Scarcity}: Despite the abundance of medical images, computational pathology is still data-hungry for specific tasks \cite{chen2022fast,le2025moma}. The high cost of image acquisition and annotation cost necessitates the development of more efficient and effective methods.

To tackle these challenges, existing methods primarily adopt multiple-instance learning (MIL) and fine-tuning strategies \cite{lee2025benchmarking}. MIL approaches partition WSIs into a disjoint set of patches, extract image embeddings from these patches, and aggregate these embeddings to produce the final output. Despite the immense size of WSIs, MIL models have proved their effectiveness in many applications \cite{clam,abmil,transmil,dtfd}.
Fine-tuning approaches, on the other hand, aim to develop task-specific models by retraining or fine-tuning existing models that are previously trained on large-scale datasets. When applied to multiple tasks, these methods typically develop new models independently for each task. Though this can reduce computational costs, it prevents interaction between models, limiting their potential synergies and improvements. Moreover, naively fine-tuning on a new task without careful optimization can lead to catastrophic forgetting, where the model loses previously acquired knowledge during training on a new task \cite{li2019learn,perkonigg2021dynamic}. To mitigate this issue, extensive research in continual learning (CL) \cite{agem,icarl,derpp} has focused on developing techniques that allow models to learn from new tasks while preserving prior knowledge. However, adapting CL to WSIs requires further investigation, since tasks are distinct from one another, yet tissue characteristics often exhibit similarities across organs (\textit{e.g.}, breast and prostate, or esophagus and cervix). Furthermore, the limited number of samples in WSI datasets make naive CL adaptations prone to producing ambiguous representations, ultimately deteriorating model robustness and generalization ability.

\begin{figure}[http]
\centerline{\includegraphics[width=1\columnwidth]{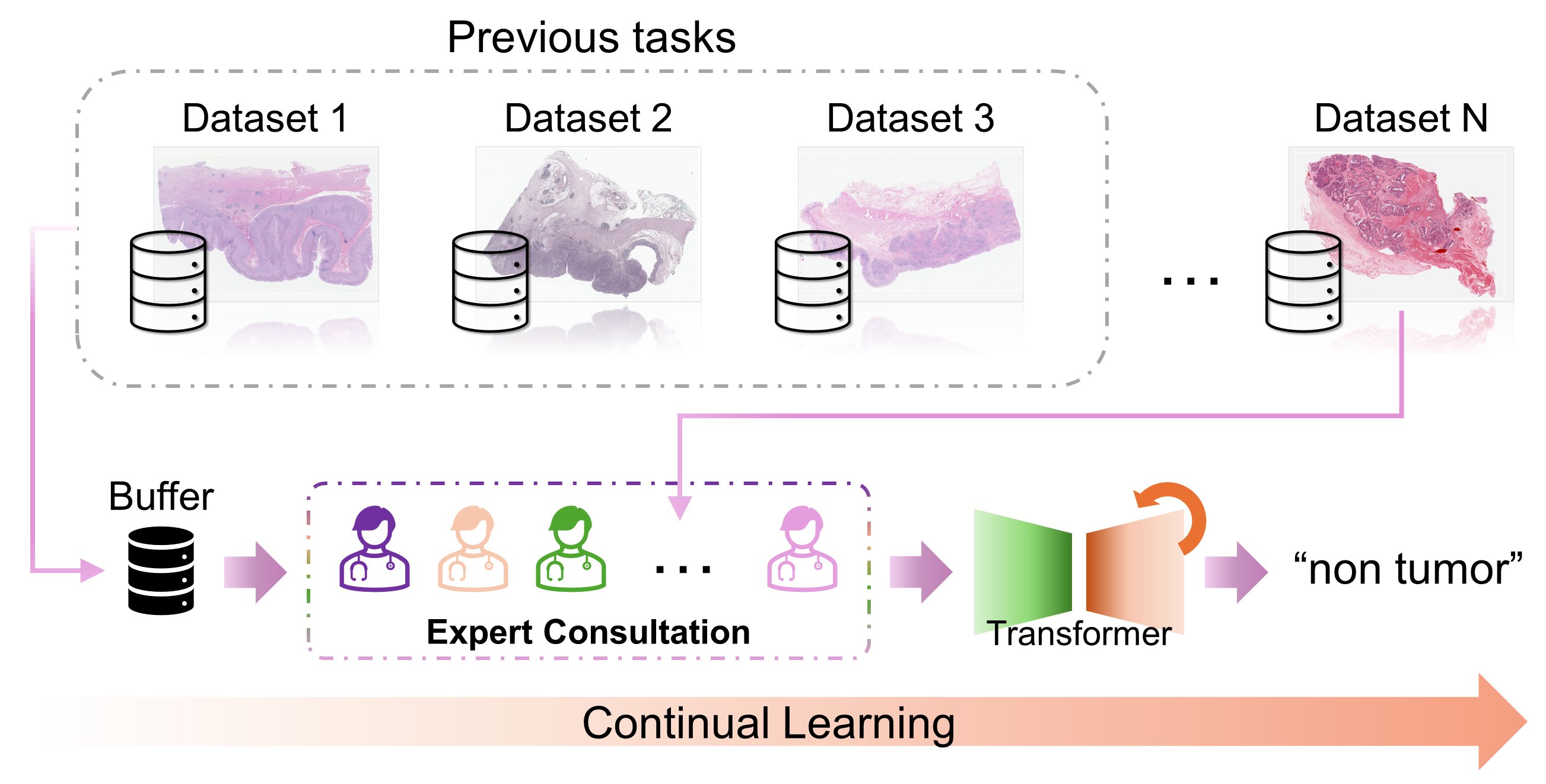}}
\caption{Illustration of COSFormer in continual learning manner.}
\label{fig:thumbnail}
\end{figure}

In this study, we introduce COSFormer, a \textbf{CO}ntinual learning approach for whole \textbf{S}lide image analysis with a Trans\textbf{Former} architecture. An overview of COSEFormer is illustrated in Fig. \ref{fig:thumbnail}. COSFormer introduces an expert consultation mechanism combined with Transformer-based autoregressive decoding for knowledge aggregation and decision-making. Unlike traditional approaches, COSFormer utilizes a single core model to handle multiple WSI analysis tasks in a continuous and sequential fashion, which can reduce model storage and computation cost without compromising model performance. 
To avoid task ambiguity in the model space caused by learning multiple tasks, we develop an expert consultation module that facilitates the learning of effective, task-relevant embeddings throughout the continuous learning process. This module is inspired by real-life medical diagnostic consultations and is implemented using a mixture-of-experts (MoEs) technique. For decision-making or classification, instead of using traditional multi-layer perceptrons (MLPs) or linear layers, we adapt Transformer-based autoregressive decoding from language models to produce classification outputs as diagnostic terms. This allows the model to seamlessly handle multiple distinct tasks with no or minimal modifications while enhancing performance. 
Last but not least, COSFormer employs CL to extend to new tasks without the need for retraining on previous datasets, making it more adaptable, resource-efficient, and practical for clinical deployment. 
COSFormer follows a rehearsal-based approach, where a small number of past WSIs is stored in a buffer and replayed during training on new tasks. To maximize both efficiency and effectiveness, COSFormer introduces two key strategies: 1) Text-based buffer storage: By integrating visual and text encoders, COSFormer constructs a compact yet diverse and comprehensive buffer for improved knowledge retention; 2) Past-to-present learning strategy: During training, COSFormer aligns the predicted logits from previous tasks with those generated for current tasks, effectively mitigating catastrophic forgetting and ensuring continuity in knowledge acquisition.

To demonstrate the effectiveness of COSFormer, we introduce a benchmark comprising seven WSI datasets: CAMELYON16, TCGA-NSCLC, TCGA-BRCA, TCGA-RCC, TCGA-ESCA, TCGA-TGCT, and TCGA-CESC. These datasets are designed to address tasks such as breast tumor classification (CAMELYON16) and cancer subtyping in lung (TCGA-NSCLC), breast (TCGA-BRCA), kidney (TCGA-RCC), esophagus (TCGA-ESCA), testis (TCGA-TGCT), and uterus (TCGA-CESC). COSFormer and other CL methods are sequentially trained and tested on these datasets.

The key contributions of this study are as follows:

\begin{itemize}
    \item We propose \textbf{COSFormer}, a unified \textit{Transformer-based model} designed for continuous and seamless learning across multiple WSI analysis tasks. It integrates \textit{Expert Consultation} via a mixture-of-experts mechanism to generate task-specific embeddings and employs \textit{Autoregressive Decoding} for accurate and flexible diagnostic classification.
    
    \item We introduce an effective continual learning strategy that incorporates a \textit{Text-based Buffer Storage} and \textit{Past-to-Present Learning}, enabling streamlined adaptation to new tasks without retraining on the full dataset.
    
    \item We establish a comprehensive benchmark comprising seven WSI datasets, including CAMELYON16, TCGA-NSCLC, TCGA-BRCA, TCGA-RCC, TCGA-ESCA, TCGA-TGCT, and TCGA-CESC, and demonstrate that COSFormer consistently outperforms existing continual learning methods on this benchmark.
\end{itemize}

\section{Related Works}

\noindent\textbf{Mutiple Instance Learning for WSIs.} Recent advances in MIL have substantially improved WSI analysis across various tasks. Early works, such as ABMIL \cite{abmil}, used simple attention mechanisms to identify key patches, while CLAM \cite{clam} refined this by pinpointing both the most and least contributory patches. TransMIL \cite{transmil} was the first to integrate Vision Transformer (ViT) concepts into MIL for WSIs, exploring long-range dependencies among patches. To address the challenge of limited sample sizes, DTFD-MIL \cite{dtfd} introduced a multi-stage approach with a sub-MIL branch enhancing patch embeddings before a global-MIL aggregation for final predictions. More recently, FALFormer \cite{falformer} further improved WSI analysis by redefining the landmark selection process in Nystr\"om self-attention \cite{nystrom}.

\noindent\textbf{Mixture-of-Expert in Multi-Task Learning.} Several methods have been proposed to aggregate knowledge from multiple experts for multi-task learning. 
For instance, \cite{taskexpert} proposed decomposing feature maps into representative features, each processed by an expert using an MLP. These are then aggregated into task-specific feature maps via a gating mechanism, with a memory bank enhancing long-range dependencies during training.
\cite{chen2023adamv} incorporated a sparse MoE layer into the ViT backbone \cite{vit}, featuring router networks for each task along with multiple MLP experts. Each router network selects the number of experts to engage and determines their contributions to the task-specific representation.
In a different approach, assuming only pre-trained models are available, \cite{taskmerging} utilized task vectors \cite{taskvector} and a learnable router to generate weights for knowledge retrieval, integrating them with a base model like CLIP-ViT-B/32 \cite{clip} to better align features across multiple tasks.

\noindent\textbf{Continual Learning Frameworks.} There have been various continual learning approaches applied to image classification tasks. For instance, GDumb \cite{gdumb} prevented forgetting by retraining a new model using a balanced memory buffer. ER-ACE \cite{erace} reweighted losses during experience replay to balance new and previous tasks. A-GEM \cite{agem} prevented interference among tasks by constraining gradients from new tasks. DER++ \cite{derpp} enhanced replay by using distillation to align current predicted logits with past logits.

\section{Methodology}

\begin{figure*}[http]
\centerline{\includegraphics[width=1\textwidth]{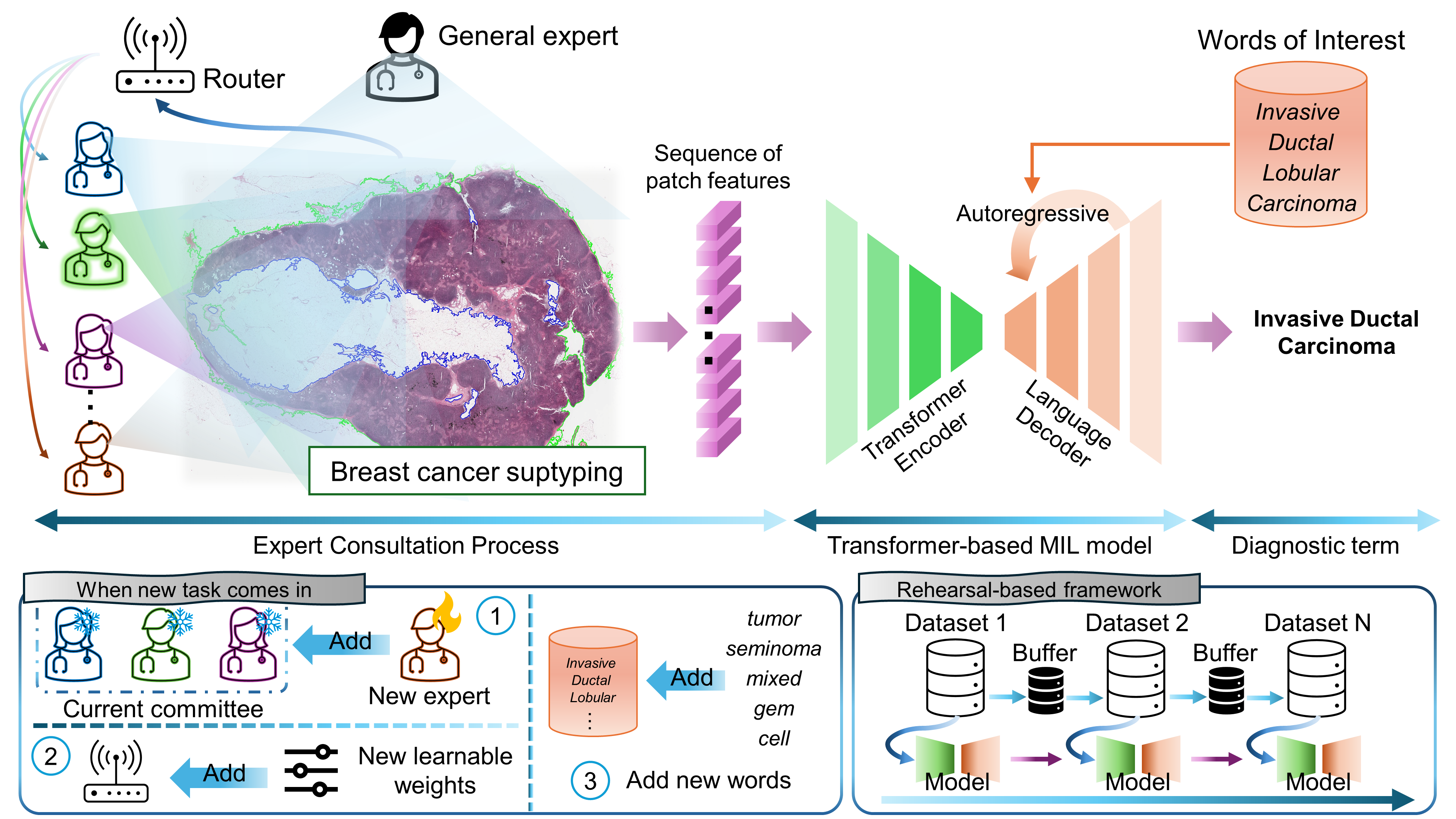}}
\caption{Overview of COSFormer. The Expert Consultation $EC$ module mimics doctor consultations, aggregating knowledge from $T$ experts via a router $\mathcal{R}$ to generate patch embeddings. A Transformer Encoder $\mathcal{E}$ processes these embeddings, followed by a Transformer Decoder with Autoregressive Decoding $\mathcal{D}$ for classification. For new tasks, a learnable expert $\theta_i$ joins, router weights update, and vocabulary $\mathcal{V}$ expands if needed. Representative slides from past tasks are retrieved using visual and text encoders and stored in the buffer $\mathcal{B}_r$ for replay.}
\label{fig:overview}
\end{figure*}

\subsection{Problem Statement}

\noindent\textbf{Problem Definition.} Let $\mathbf{D} = \{\mathcal{D}_t\}_{t=1}^{N_T}$ be the datasets for $N_T$ tasks. Our goal is to develop a model $\mathcal{F}$ that can sequentially learn these $N_T$ tasks on $\mathbf{D}$. Each task varies, such as cancer classification or cancer subtyping for a WSI, and contains $c_i$ categories to be classified. When training on the $t$-th task, the model $\mathcal{F}$ is expected to retain knowledge from the past tasks, \textit{i.e.}, from the $1$-st to the $(t-1)$-th tasks, without forgetting prior knowledge. For inference, we evaluate $\mathcal{F}$ under two scenarios: class-incremental learning (CLASS-IL) and task-incremental learning (TASK-IL).

\noindent\textbf{Class-incremental Learning.} The model $\mathcal{F}$ is blind to the target task. Trained on $N_T$ tasks, $\mathcal{F}$ produces logits over a cumulative set of $N_c$ classes, where $N_c = \sum_{i=1}^{N_T} c_i$ and $c_i$ denotes the number of classes in task $i$. $\mathcal{F}$ must distinguish among all previously seen classes without any task-specific information.

\noindent\textbf{Task-incremental Learning.} The model $\mathcal{F}$ is aware of the current target task $\mathcal{T}$ and its class boundary. In other words, it generates logits only for the classes relevant to $\mathcal{T}$ since it knows which task it is targeting and the specific classes of $\mathcal{T}$.


\subsection{COSFormer Architecture}
\noindent \textbf{Overview.} We design COSFormer as the model $\mathcal{F}$, composed of three key components: 1) \textit{Expert Consultation} module: It converts patch embeddings into the model space; 2) \textit{Transformer-based backbone}: It captures long-range dependencies among patch and word embeddings; 3) \textit{Autoregressive Decoding}: It iteratively predicts labels word-by-word, where each word is selected from a vocabulary $\mathcal{V}$.

\begin{figure}[http]
\centerline{\includegraphics[width=1.0\columnwidth]{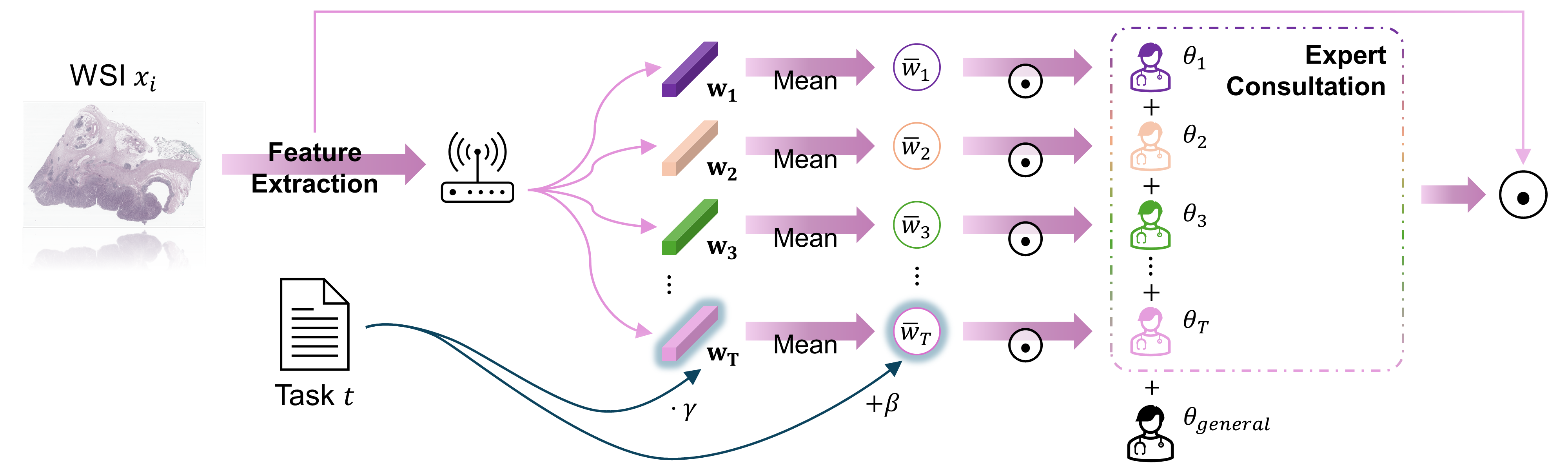}}
\caption{Illustration of Expert Consultation module ($EC$).}
\label{fig:ec}
\end{figure}

\noindent \textbf{Expert Consultation.} Following \cite{bui2024mecformer}, we built Expert Consultation module as shown in Fig. \ref{fig:ec}. In continual learning with $N_T$ tasks, we establish a generalist $\theta_{general}$ and an expert committee $\boldsymbol{\theta} = \{\theta_i\}_{i=1}^{N_T}$ for $N_T$ tasks, where each expert $\theta_i \in \mathbb{R}^{d_f \times d_{model}}$ is assigned to the $i$-th task, both $\theta_{general}$ and $\theta_i$ contain trainable parameters, and $d_f$ and $d_{model}$ are the embedding sizes of the frozen feature extractor and the Transformer-based backbone, respectively. The generalist $\theta_{general}$ oversees all tasks, while each expert $\theta_i$ learns its assigned task.
Given a sequence of $N$ patch embeddings $\mathbf{z} = \{z_i\}^{N}_{i=1}$ from a WSI at the target task $\mathcal{T}$, we define a router $\mathcal{R}$ to generate weights for $N_T$ tasks, denoted as $\mathcal{W} = \{\mathbf{w}_i\}_{i=1}^{N_T} = FC_2\big(\texttt{ReLU}(FC_1(\mathbf{z})\big)$, where $FC_1$ and $FC_2$ are two linear layers and $\mathbf{w}_i \in \mathbb{R}^{N}$ is a weight vector for all $N$ patches in the WSI. These weight vectors are adjusted to lean towards the target task $\mathcal{T}$ as follows:

\begin{equation}
    \tilde{\mathbf{w}}_i = \frac{\exp\big(\mathbbm{1}_{i=\mathcal{T}}(\mathbf{w}_i \cdot \mathbf{1}\gamma) + \mathbbm{1}_{i\neq \mathcal{T}}(\mathbf{w}_i)\big)}{\sum^{N_T}_{i \neq \mathcal{T}} \exp(\mathbf{w}_i) + \mathbf{w}_t \cdot \mathbf{1}\gamma },
    \label{eq:1}
\end{equation}

\noindent where $\gamma$ is a scaling hyper-parameter. Averaging $\mathbf{\tilde{w}}_i$, we obtain a scalar weight $\tilde{w}_i$ for each task. We then further refine these weights by shifting the target weight. The procedure can be formulated as follows:

\begin{equation}
 \bar{w}_i = \frac{1}{N} \sum^{N}_{j=1} \tilde{w}_{ij} + \mathbbm{1}_{i=\mathcal{T}} \cdot \beta,
\label{eq:2}
\end{equation}

\noindent where $\beta$ is a shifting hyper-parameter. Using $\bar{\mathbf{w}} = \{\bar{w}_i\}^{N_T}_{i=1}$, we aggregate all experts $\boldsymbol{\theta}=\{\theta_i\}^{N_t}_{i=1}$ and incorporate them into the generalist $\theta_{general}$, yielding the expert consultation in the form of projection weights $\theta_{EC}$, given by:

\begin{equation}
\begin{array}{ll}
\theta_{EC} = \theta_{general} + \sum^{N_T}_{i=1} \theta_i \cdot \bar{w}_i.
\end{array}
\label{eq:3}
\end{equation}

\noindent $\theta_{EC}$ serves as projection weights to transform $\mathbf{z} \in \mathbb{R}^{d_f}$ into $\mathbf{z}' = \mathbf{z} \cdot \theta_{EC} \in \mathbb{R}^{d_{model}}$. We denote Eqs. \eqref{eq:1}, \eqref{eq:2}, and \eqref{eq:3} as the operator $EC(\cdot)$ for convenience.

\noindent \textbf{Transformer-Based Backbone with Autoregressive Decoding.} 
We adopt an encoder-decoder architecture with autoregressive decoding for classification, formulated as:

\begin{equation}
\hat{p}^{(k+1)} = \mathcal{D}\big(\mathcal{E}(\mathbf{z}') \mid \hat{p}^{(k)}\big),
\label{eq:4}
\end{equation}

\noindent where $\mathbf{z}' = EC(\mathbf{z}, \mathcal{T})$ and $\mathcal{E}$ and $\mathcal{D}$ denote the Transformer encoder and decoder, respectively. $\mathcal{E}$ encodes the output of $EC$, and $\mathcal{D}$ uses the output of $\mathcal{E}$ along with the previous prediction $\hat{p}^{(k)}$ to generate $\hat{p}^{(k+1)}$. Notably, $\hat{p}^{(k)}$ is mapped to the pre-extracted word embedding from a text encoder for compatibility with $\mathcal{D}$. The logit vector $\hat{p}^{(k)} \in \mathbb{R}^{N_{\mathrm{voc}}}$ represents the probabilities for all words in the vocabulary $\mathcal{V}$, with the predicted word selected as 
$\hat{y}^{(k)} = \mathcal{V}(u^{*})$ where $u^{*} = \arg\max_{u}(\hat{p}^{(k)}_u)$. The sequence begins with $\hat{y}^{(0)} = \texttt{<BOS>}$ and terminates when $\hat{y}^{(k+1)} = \texttt{<EOS>}$.

\noindent \textbf{Encoder.} The encoder $\mathcal{E}$ consists of $N_e$ layers, each using Nystr\"om self-attention (NA) \cite{nystrom} to process WSI patches, followed by a normalization layer (Norm) and a skip connection:
\begin{equation}
\mathbf{z}_{i+1} = \mathbf{z}_i + \text{Norm}\big(\text{NA}(\mathbf{z}_i)\big), \quad i = 1, \dots, N_e,
\label{eq:encoder}
\end{equation}

\noindent where $\mathbf{z}_0 = \mathbf{z}'$.

\noindent \textbf{Decoder.} The decoder $\mathcal{D}$ is composed of $N_d$ layers. Each layer utilizes two multi-head self-attention (MHSA) blocks. The first MHSA processes previous word embeddings, followed by Norm and a skip connection. The second MHSA performs cross-attention between $\mathcal{E}$'s patch embeddings and previous word embeddings, followed by Norm, a skip connection, and a feed-forward network (FFN), expressed as:


\begin{equation}
\begin{aligned}
\mathbf{h}_{i} &= F_{\text{text}}\big(\{\mathcal{V} (\arg\max(p^{(k)}))\}_{k=1}^{K}\big), \\
\mathbf{h}'_{i} &= \mathbf{h}_{i} + \text{Norm}\big(\text{MHSA}(\mathbf{h}_{i}, \mathbf{h}_{i}, \mathbf{h}_{i})\big), \\
\mathbf{h}''_{i} &= \text{FFN}\Big(\mathbf{h}'_{i} + \text{Norm}\big(\text{MHSA}(\mathbf{z}_{N_e}, \mathbf{h}'_{i}, \mathbf{h}'_{i})\big)\Big), \\ 
i &= 1, \dots, N_d,
\end{aligned}
\label{eq:decoder2}
\end{equation}

\noindent where $K$ denotes the number of decoding steps taken so far and $F_{\text{text}}$ represents the text encoder for word embedding. After passing through $N_d$ decoder layers, the next-word logit prediction is given by:
\begin{equation}
p^{(k+1)} = \text{Linear}(\mathbf{h}''_{N_d}),
\label{eq:linear}
\end{equation}

\noindent where $\text{Linear}$ projects $\mathbf{h}''_{N_d} \in \mathbb{R}^{d_{\text{model}}}$ to $p^{(k+1)} \in \mathbb{R}^{N_{\mathrm{voc}}}$.

\noindent \textbf{Inference.} 
Inference is performed under two scenarios: TASK-IL and CLASS-IL, each following a distinct procedure. In the TASK-IL scenario, the model is aware of the target task $\mathcal{T}$ with its associated class labels, and thus it knows the corresponding Words of Interest, denoted as $WoI_{\mathcal{T}}$, in the vocabulary $\mathcal{V}$. Using $WoI_{\mathcal{T}}$, all irrelevant words are masked out:
\begin{equation}
  \hat{p}^{(k)}_{j} :=
    \begin{cases}
      \hat{p}^{(k)}_{j} & \text{if \(w_j \in {WoI}_{\mathcal{T}}\)}, \\
      -\infty & \text{otherwise},
    \end{cases}
\label{eq:mask}
\end{equation}

\noindent where $w_j$ is the $j$-th word in the vocabulary $\mathcal{V}$. 
In contrast, in the CLASS-IL scenario, the model is not informed of the target task \(\mathcal{T}\), meaning that all words are considered relevant and no masking is applied to the logit prediction. Accordingly, during training, we set $\gamma = 1$ and $\beta = 0$. As a result, the weights are not updated in Eqs.~\eqref{eq:1} and \eqref{eq:2}.

\subsection{Continual Learning Training} \label{sec:continual-learning}
\noindent \textbf{New Task Definition.} For continual learning, we start with an empty expert committee, $\boldsymbol{\theta} = \varnothing$, and an empty vocabulary, $\mathcal{V} = \varnothing$.
When a new task arrives, the model undergoes the following three steps:
1) Define a new learnable matrix, $\theta_i \in \mathbb{R}^{d_f \times d_{model}}$;
2) Define a new learnable projection weight in the router, $\mathcal{R}$;
3) Add new words to the vocabulary $\mathcal{V}$ and new learnable parameters for the classification head in the Transformer Decoder $\mathcal{D}$ if necessary.
Notably, during training on the $(t+1)$-th task, experts from previous tasks, $\boldsymbol{\theta} = \{\theta_i\}_{i\leq t}$, are frozen, and only the new expert $\theta_{t+1}$, and the generalist $\theta_{general}$ remain learnable.

\begin{figure}[http]
\centerline{\includegraphics[width=1\columnwidth]{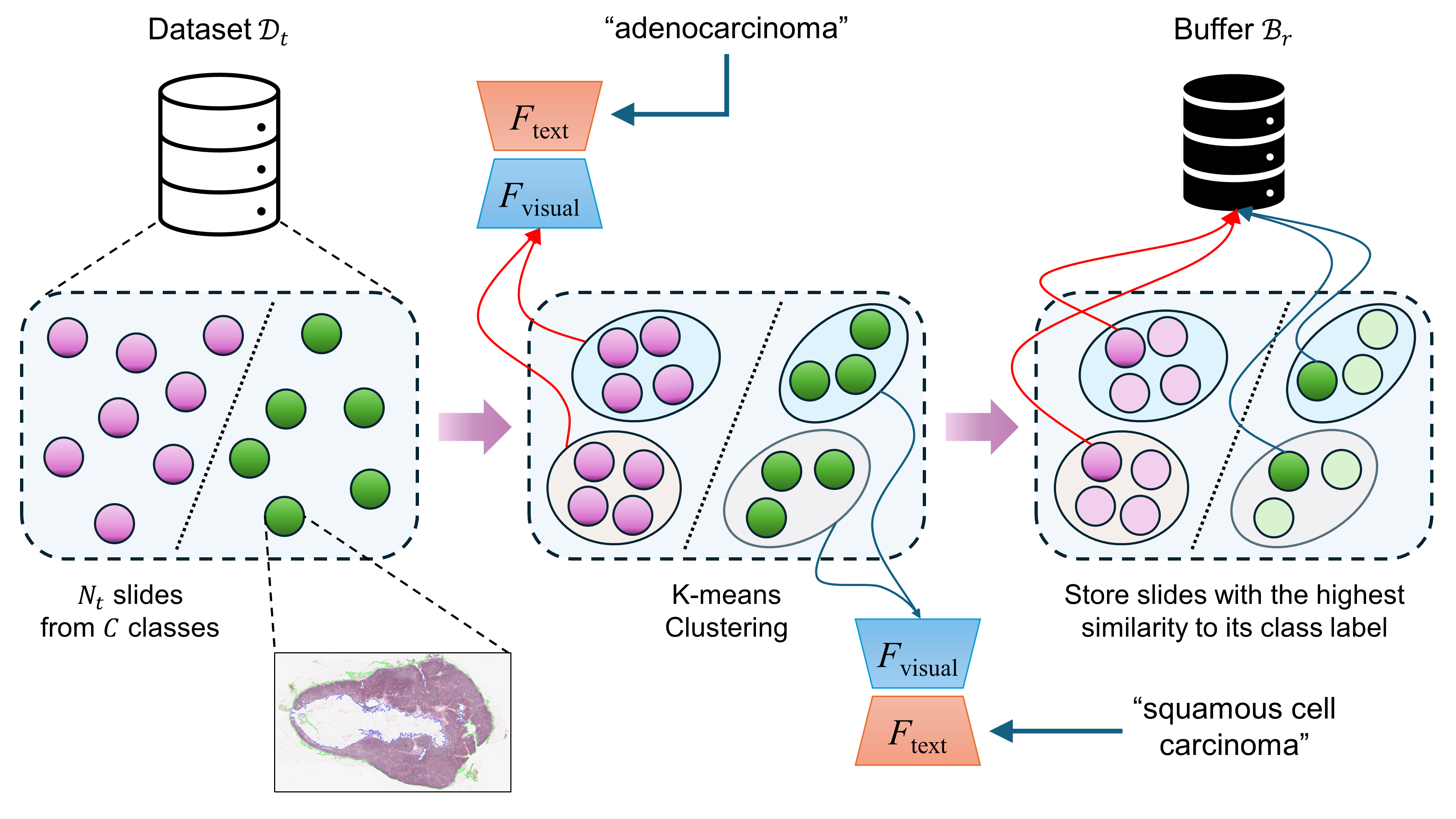}}
\caption{Illustration of Text-based Retrieval Buffer Storage Strategy.}
\label{fig:retrieval}
\end{figure}

\noindent \textbf{Text-based Retrieval Buffer Storage Strategy.} In continual learning, we adopt a rehearsal-based approach by initializing an empty buffer, \(\mathcal{B}_r = \varnothing\). When training on the $(t+1)$-th task, a subset of representative samples from task $t$ is stored using \emph{Text-based Retrieval Buffer Storage Strategy} (Fig. \ref{fig:retrieval}). This strategy leverages a visual encoder \(F_{\text{visual}}\) and a text encoder \(F_{\text{text}}\) to identify most informative samples. Let the WSI dataset for the $t$-th task be $\mathcal{D}_t = \{(x_i, y_i)\}_{i=1}^{N_t}$ where $N_t$ is the total number of WSIs and \(y_i \in \{1,\dots,C_t\}\) denotes the class label of a WSI \(x_i\). 
To facilitate structured and efficient sampling, we first group the WSIs according to their class labels:
\begin{equation}
\mathcal{D}_t^c = \{x_i \in \mathcal{D}_t \mid y_i = c\}, \quad c = 1,\dots, C_t.
\label{eq:5}
\end{equation}

\noindent For each WSI \(x_i \in \mathcal{D}_t^c\), we extract patch embeddings using the visual encoder \(F_{\text{visual}}\):
\begin{equation}
\mathbf{z}^v_i = F_{\text{visual}}(x_i) = \{z^v_{i,j}\}_{j=1}^{N_i},
\label{eq:visual_features}
\end{equation}
where $\mathbf{z}^v_i$ is a set of $N_i$ patch embeddings from $x_i$. Next, we encode the name of class $c$ (e.g., "adenocarcinoma"), denoted as $\operatorname{Text}(c)$, using the text encoder $F_{\text{text}}$:
\begin{equation}
z^t_c = F_{\text{text}}(\operatorname{Text}(c)).
\label{eq:text_encoding}
\end{equation}

\noindent We then compute the importance scores $\mathbf{S}^c_{t}$ for the $t$-th task and class $c$ as follows:
\begin{equation}
\begin{aligned}
\mathbf{S}^c_t &= \{ s_i \mid x_i \in \mathcal{D}_t^c \}, \\
s_i &= \max_{1 \leq j \leq N_i} \Bigl( \langle z^v_{i,j}, {z^t_c}^\intercal \rangle \Bigr)
\label{eq:8}
\end{aligned}
\end{equation}

\noindent where $s_i$ is the importance score for a WSI $x_i$ and $\langle z^v_{i,j}, {z^t_c}^\intercal \rangle$ denotes the similarity score (\textit{e.g.}, dot product) between the patch embedding \(z^v_{i,j}\) and the class embedding \(z^t_c\).  


To ensure diversity among the selected WSIs, we apply $k$-means clustering to the WSIs in $\mathcal{D}_t^c$ to form $K$ clusters based on Euclidean distances computed from the set of patch embeddings \(\mathbf{z}^v_i\). The $k$-th cluster is denoted as $\mathcal{D}_t^{c,k}$, with its corresponding importance scores forming the set $\mathbf{S}_t^{c,k}$. From each cluster, we select the WSI with the highest importance score with respect to the class embedding and add the selected WSI to the rehearsal buffer as follows:
\begin{equation}
\mathcal{B}_r \leftarrow \mathcal{B}_r \cup \Bigl\{ \arg\max_{x_i} \mathbf{S}_t^{c,k} (x_i) \Bigr\}, \quad k = 1,\dots, K.
\label{eq:9}
\end{equation}

\noindent Equation \eqref{eq:9} is applied for each class $c = 1, \dots, C_t$ to complete the selection of representative WSIs from the dataset \(\mathcal{D}_t\). For fair comparison with other continual learning methods, we constrain the size of the buffer to at most $N_{\mathrm{buf}}$ WSIs. If $|\mathcal{B}_r| > N_{\mathrm{buf}}$, we randomly delete WSIs from $\mathcal{B}_r$ until $|\mathcal{B}_r| = N_{\mathrm{buf}}$.

\noindent \textbf{Past-to-Present Learning.} During training on the current $t$-th task, we employ three learning objectives: 1) supervised learning on samples from the current task $t$; 2) supervised learning on samples from past tasks retrieved from the buffer $\mathcal{B}_r$; 3) dark experience replay on past samples, inspired by \cite{derpp}. The final loss function for learning on a WSI $x_i$ at the $k$-th decoding step, produced by the Transformer Decoder $\mathcal{D}$, is formulated as follows:

\begin{equation}
\begin{array}{ll}
\mathcal{L}_{\text{CE}}\Bigl(p_{t,i}^{(k)},\, y_i^{(k)} \mid x_i \in \mathcal{D}_t\Bigr)
+ \mathcal{L}_{\text{CE}}\Bigl(p_{t,i}^{(k)},\, y_i^{(k)} \mid x_i \in \mathcal{B}_r\Bigr) \\
+ \mathcal{L}_{\text{MSE}}\Bigl(p_{t,i}^{(k)},\, p_{t-1,i}^{(k)} \mid x_i \in \mathcal{B}_r\Bigr),
\end{array}
\end{equation}

\noindent where $p^{(k)}_{t,i} \in \mathbb{R}^{N_{\text{voc}}}$ denotes the predicted logits at the $k$-th decoding step for WSI $x_i$, produced by the model trained on the $t$-th task, and $y^{(k)}_i$ is the corresponding ground-truth. $\mathcal{L}_{CE}$ and $\mathcal{L}_{MSE}$ represent the cross-entropy loss and the mean squared error loss, respectively. The first two terms perform supervised classification on both current samples and samples retrieved from the buffer, while the third term minimizes the difference between current and past model predictions to mitigate catastrophic forgetting.

\section{Experiments}

\begin{table*}[!ht]
\centering
\caption{Results of COSFormer and other CL methods on a sequence of tasks from CAMELYON16 to TCGA-CESC (Metric: Accuracy (\%) $\pm$ Standard Deviation).}
\resizebox{\textwidth}{!}{
\begin{tabular}{l|c|c|c|c|c}
\toprule
\textbf{Dataset} & \textbf{GDumb} & \textbf{ER-ACE} & \textbf{A-GEM} & \textbf{DER++} & \textbf{COSFormer (Ours)} \\
\midrule
\multicolumn{6}{l}{\textit{Task-incremental Learning (TASK-IL)}} \\
\midrule
CAMELYON16 & 49.871 ($\pm$16.377) & \textbf{94.057 ($\pm$5.714)} & 78.811 ($\pm$22.632) & 91.473 ($\pm$5.853) & \underline{93.282 ($\pm$2.368)} \\
TCGA-NSCLC & 47.908 ($\pm$8.518) & 81.240 ($\pm$5.680) & 87.896 ($\pm$5.521) & \textbf{91.372 ($\pm$2.873)} & \underline{89.754 ($\pm$2.931)} \\
TCGA-BRCA & 51.754 ($\pm$25.879) & 84.931 ($\pm$3.416) & 85.669 ($\pm$6.389) & \underline{92.433 ($\pm$0.201)} & \textbf{93.700 ($\pm$1.215)} \\
TCGA-RCC & 43.901 ($\pm$16.858) & 90.372 ($\pm$3.576) & 94.240 ($\pm$3.026) & \underline{94.614 ($\pm$2.897)} & \textbf{95.010 ($\pm$2.880)} \\
TCGA-ESCA & 45.541 ($\pm$4.980) & 67.792 ($\pm$19.207) & \textbf{92.814 ($\pm$1.787)} & \underline{91.006 ($\pm$4.763)} & 89.773 ($\pm$8.209) \\
TCGA-TGCT & 30.194 ($\pm$9.064) & 77.935 ($\pm$7.179) & \underline{89.088 ($\pm$7.542)} & 89.056 ($\pm$4.471) & \textbf{95.123 ($\pm$4.406)} \\
TCGA-CESC & 37.850 ($\pm$6.249) & 84.652 ($\pm$1.173) & 94.062 ($\pm$2.306) & \underline{95.309 ($\pm$3.222)} & \textbf{95.317 ($\pm$1.557)} \\
\rowcolor[HTML]{DCDCDC}
\textbf{Average} & 43.860 ($\pm$6.506) & 82.997 ($\pm$2.709) & 88.940 ($\pm$2.264) & \underline{92.180 ($\pm$1.986)} & \textbf{93.137 ($\pm$2.700)} \\
\midrule
\multicolumn{6}{l}{\textit{Class-incremental Learning (CLASS-IL)}} \\
\midrule
CAMELYON16 & 7.235 ($\pm$10.525) & \textbf{91.731 ($\pm$8.360)} & 72.351 ($\pm$17.165) & \underline{87.597 ($\pm$7.634)} & 83.462 ($\pm$5.608) \\
TCGA-NSCLC & 4.271 ($\pm$4.315) & \underline{76.940 ($\pm$4.938)} & 30.155 ($\pm$4.844) & 70.612 ($\pm$10.381) & \textbf{89.073 ($\pm$3.489)} \\
TCGA-BRCA & 2.122 ($\pm$1.971) & \underline{80.331 ($\pm$4.664)} & 67.019 ($\pm$12.241) & \textbf{89.927 ($\pm$1.093)} & 55.347 ($\pm$24.508) \\
TCGA-RCC & 3.831 ($\pm$3.511) & 88.069 ($\pm$2.458) & \underline{92.698 ($\pm$1.727)} & \textbf{93.838 ($\pm$3.345)} & 78.437 ($\pm$15.921) \\
TCGA-ESCA & 10.757 ($\pm$6.149) & \underline{43.907 ($\pm$30.411)} & 21.039 ($\pm$11.832) & 27.479 ($\pm$15.079) & \textbf{90.996 ($\pm$4.812)} \\
TCGA-TGCT & 0.000 ($\pm$0.000) & 32.161 ($\pm$32.519) & 18.166 ($\pm$11.086) & \underline{34.552 ($\pm$28.967)} & \textbf{75.967 ($\pm$1.087)} \\
TCGA-CESC & 0.623 ($\pm$1.079) & 74.248 ($\pm$5.106) & 93.742 ($\pm$1.902) & \textbf{94.383 ($\pm$2.764)} & \underline{94.350 ($\pm$1.889)} \\
\rowcolor[HTML]{DCDCDC}
\textbf{Average} & 4.120 ($\pm$1.420) & 69.627 ($\pm$9.553) & 56.453 ($\pm$3.631) & 71.198 ($\pm$6.308) & \textbf{81.090 ($\pm$4.649)} \\
\bottomrule
\end{tabular}}
\label{tab:main_results}
\end{table*}

\begin{table*}[!ht]
\centering
\caption{Results of COSFormer and other CL methods on a reversed sequence of tasks from TCGA-CESC to CAMELYON16 (Metric: Accuracy (\%) $\pm$ Standard Deviation).}
\resizebox{\textwidth}{!}{
\begin{tabular}{l|c|c|c|c|c}
\toprule
\textbf{Dataset} & \textbf{GDumb} & \textbf{ER-ACE} & \textbf{A-GEM} & \textbf{DER++} & \textbf{COSFormer (Ours)} \\
\midrule
\multicolumn{6}{l}{\textit{Task-incremental Learning (TASK-IL)}} \\
\midrule
TCGA-CESC & 64.834 ($\pm$24.111) & 93.774 ($\pm$3.689) & 90.305 ($\pm$1.783) & \textbf{94.039 ($\pm$2.360)} & \underline{94.012 ($\pm$2.497)} \\
TCGA-TGCT & 43.215 ($\pm$6.348) & 79.966 ($\pm$19.562) & 33.751 ($\pm$16.690) & \textbf{89.146 ($\pm$6.708)} & \underline{87.983 ($\pm$0.543)} \\
TCGA-ESCA & 51.450 ($\pm$6.248) & 87.392 ($\pm$4.909) & 83.149 ($\pm$12.208) & \underline{87.987 ($\pm$6.426)} & \textbf{88.582 ($\pm$5.925)} \\
TCGA-RCC & 29.551 ($\pm$15.117) & 91.540 ($\pm$4.033) & \underline{91.936 ($\pm$3.011)} & 91.170 ($\pm$4.621) & \textbf{93.861 ($\pm$3.497)} \\
TCGA-BRCA & 43.588 ($\pm$20.452) & 87.298 ($\pm$5.897) & 90.030 ($\pm$4.270) & \underline{90.675 ($\pm$4.700)} & \textbf{90.766 ($\pm$1.810)} \\
TCGA-NSCLC & 52.859 ($\pm$4.887) & 71.931 ($\pm$6.159) & \underline{92.607 ($\pm$3.575)} & 90.549 ($\pm$4.272) & \textbf{93.373 ($\pm$2.261)} \\
CAMELYON16 & 48.320 ($\pm$10.710) & 54.264 ($\pm$14.103) & \textbf{94.315 ($\pm$1.614)} & \underline{93.023 ($\pm$0.775)} & 89.406 ($\pm$2.492) \\
\rowcolor[HTML]{DCDCDC}
\textbf{Average} & 47.688 ($\pm$4.771) & 80.881 ($\pm$5.564) & 82.299 ($\pm$0.632) & \underline{90.941 ($\pm$3.156)} & \textbf{91.140 ($\pm$1.198)} \\
\midrule
\multicolumn{6}{l}{\textit{Class-incremental Learning (CLASS-IL)}} \\
\midrule
TCGA-CESC & 25.828 ($\pm$27.707) & \underline{79.311 ($\pm$4.916)} & 26.852 ($\pm$30.164) & 71.122 ($\pm$28.376) & \textbf{92.481 ($\pm$1.587)} \\
TCGA-TGCT & 0.952 ($\pm$1.347) & 67.322 ($\pm$21.422) & 33.751 ($\pm$16.690) & \textbf{85.069 ($\pm$2.329)} & \underline{82.838 ($\pm$5.297)} \\
TCGA-ESCA & 4.773 ($\pm$5.514) & \underline{75.487 ($\pm$17.828)} & 51.526 ($\pm$7.524) & 41.136 ($\pm$26.556) & \textbf{86.764 ($\pm$10.053)} \\
TCGA-RCC & 13.071 ($\pm$9.440) & \underline{79.251 ($\pm$11.898)} & 58.465 ($\pm$0.925) & \textbf{87.321 ($\pm$2.980)} & 62.657 ($\pm$9.477) \\
TCGA-BRCA & 3.356 ($\pm$1.164) & \textbf{76.728 ($\pm$9.301)} & 35.694 ($\pm$25.198) & \underline{76.002 ($\pm$2.839)} & 47.942 ($\pm$5.402) \\
TCGA-NSCLC & 1.178 ($\pm$0.939) & 68.313 ($\pm$2.851) & \textbf{91.812 ($\pm$3.865)} & 90.549 ($\pm$4.272) & \underline{91.102 ($\pm$4.642)} \\
CAMELYON16 & 13.953 ($\pm$5.626) & 42.119 ($\pm$21.050) & \textbf{94.315 ($\pm$1.614)} & \underline{93.023 ($\pm$0.775)} & 91.731 ($\pm$2.935) \\
\rowcolor[HTML]{DCDCDC}
\textbf{Average} & 9.016 ($\pm$3.729) & 69.790 ($\pm$5.293) & 56.059 ($\pm$2.919) & \underline{77.746 ($\pm$8.625)} & \textbf{79.359 ($\pm$1.502)} \\
\bottomrule
\end{tabular}}
\label{tab:main_results2}
\end{table*}

\subsection{Datasets} 
\begin{figure}[http]
\centerline{\includegraphics[width=1.0\columnwidth]{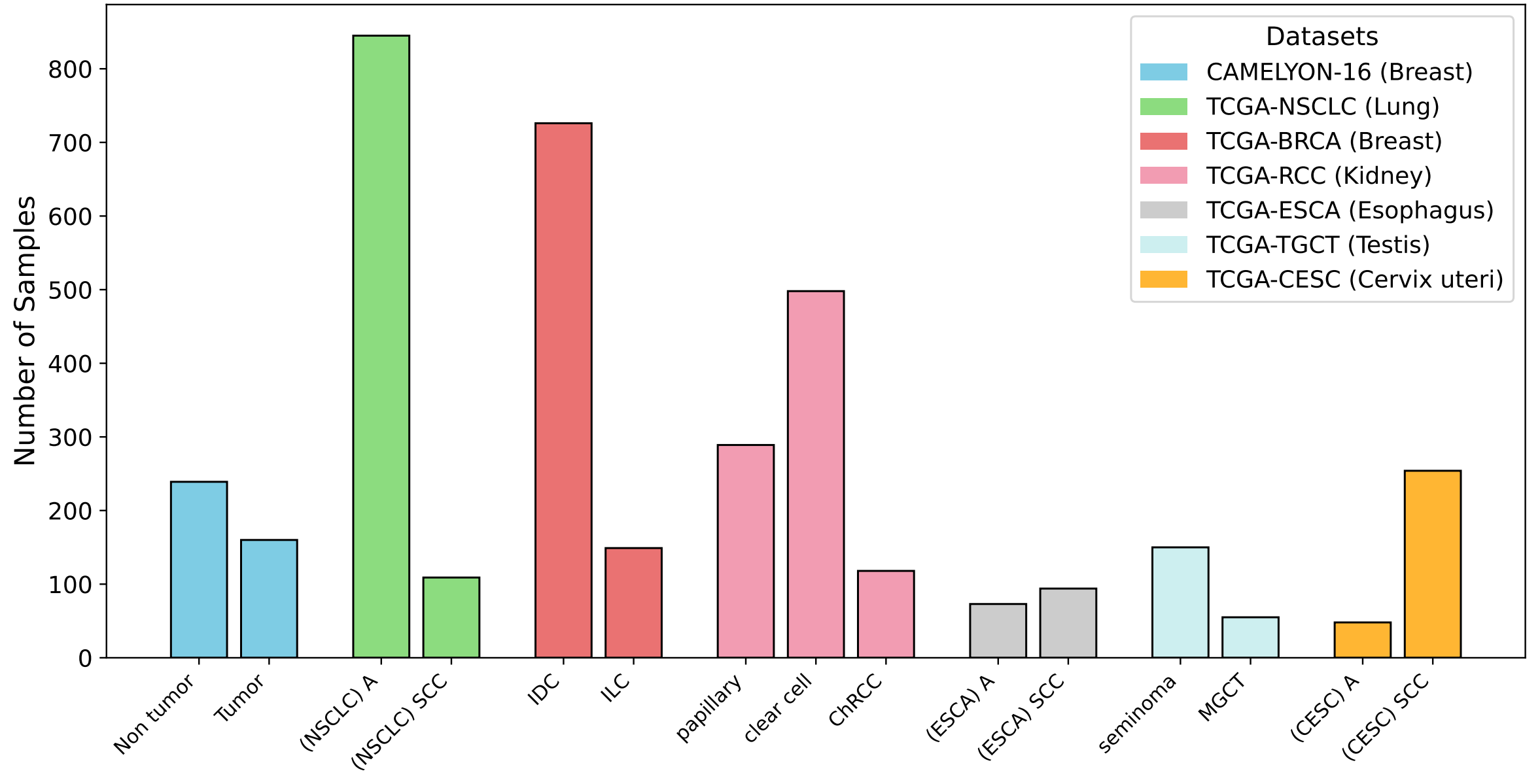}}
\caption{Distribution of seven datasets. Abbreviations: IDC, invasive ductal carcinoma; ILC, invasive lobular carcinoma; papillary, papillary carcinoma; clear cell, clear cell carcinoma; ChRCC, chromophobe renal cell carcinoma; A, adenocarcinoma; SCC, squamous cell carcinoma; S, seminoma; MGCT, mixed germ cell tumor.}
\label{fig:dataset}
\end{figure}

\noindent We construct a benchmark comprising a sequence of seven WSI datasets, covering six organs and seven tasks: CAMELYON16 (C16) \cite{camelyon} for tumor detection (tumor: 160 WSIs and non-tumor: 239 WSIs) and six cancer subtyping tasks across six organs from TCGA: TCGA-NSCLC (lung), TCGA-BRCA (breast), TCGA-RCC (kidney), TCGA-ESCA (esophagus), TCGA-TGCT (testis), and TCGA-CESC (uterus). For TCGA-NSCLC, TCGA-ESCA, and TCGA-CESC, the task is to differentiate between \textit{adenocarcinoma} (TCGA-NSCLC: 109 WSIs, TCGA-ESCA: 73 WSIs, and TCGA-CESC: 48 WSIs) and \textit{squamous cell carcinoma} (TCGA-NSCLC: 845 WSIs, TCGA-ESCA: 94 WSIs, and TCGA-CESC: 254 WSIs). TCGA-BRCA is designed to distinguish between \textit{invasive ductal carcinoma} (726 WSIs) and \textit{invasive lobular carcinoma} (149 WSIs). TCGA-RCC contains WSIs from three categories: \textit{papillary renal cell carcinoma} (289 WSIs), \textit{clear cell renal cell carcinoma} (498 WSIs), and \textit{chromophobe renal cell carcinoma} (118 WSIs). TCGA-TGCT involves two categories such as \textit{seminoma} (150 WSIs) and \textit{mixed germ cell tumor} (55 WSIs). For TCGA-NSCLC, TCGA-BRCA, and TCGA-RCC, due to the larger number of WSIs, we use a train-validation-test split of $0.8:0.1:0.1$. In contrast, for TCGA-ESCA, TCGA-TGCT, and TCGA-CESC, which have fewer slides, we adopt a split ratio of $0.48:0.12:0.4$ to maintain a balanced number of WSIs in the test set across datasets.
All WSIs from the TCGA dataset are downloaded from the GDC Data Portal\footnote{\url{https://portal.gdc.cancer.gov}}.
The statistics of these seven datasets are shown in Fig. \ref{fig:dataset}. For CAMELYON16, patch tiling is acquired at $20 \times$ magnification, while for all TCGA datasets, patch tiling is performed at $10 \times$ magnification. For all datasets, we randomly generated three different train-validation-test splits to ensure the stability of performance across methods.

\subsection{Experimental Design} \noindent All experiments were run three times using three different train-validation-test splits to ensure stability. To train COSFormer, the CL approach, described in Sec. \ref{sec:continual-learning}, was applied sequentially to the datasets in the following order: C16 $\rightarrow$ TCGA-NSCLC $\rightarrow$ TCGA-BRCA $\rightarrow$ TCGA-RCC $\rightarrow$ TCGA-ESCA $\rightarrow$ TCGA-TGCT $\rightarrow$ TCGA-CESC, which we denote as C16 $\Rightarrow$ TCGA-CESC, and its reverse order is denoted as TCGA-CESC $\Rightarrow$ C16. 
To ensure fairness in comparison, all CL methods use the same backbone with $N_e$ Transformer layers, corresponding to the Transformer Encoder $\mathcal{E}$. COSFormer utilizes Expert Consultation ($EC$) for the initial projection and a Transformer Decoder ($\mathcal{D}$) for autoregressive decoding to predict class labels. In contrast, other methods use a single linear projection before $\mathcal{E}$ and a linear classification head to produce an $N_c$-dimensional logit vector, where $N_c$ denotes the total number of categories in the dataset sequence $\mathbf{D}$ (following the same setting as in \cite{icarl}). Similar to previous CL studies, accuracy (\%) is used as the primary evaluation metric. We report the accuracies on each task after training on the last task (\textit{e.g.}, TCGA-CESC in C16 $\Rightarrow$ TCGA-CESC) as well as the average accuracy over all tasks. It is worth noting that, under the TASK-IL scenario, our model knows the target task $\mathcal{T}$, whereas this information is not provided in the CLASS-IL scenario.

\subsection{Implementation Details} \noindent Following the procedure introduced in CLAM \cite{clam}, we divide each WSI into disjoint patches. For the visual encoder $F_{\text{visual}}$, we use UNI \cite{uni}, while we adopt PubMedBERT \cite{pubmedbert} as the text encoder $F_{\text{text}}$, used in the Text-based Retrieval Buffer Storage Strategy. For Expert Consultation ($EC$), we set the scaling hyperparameter $\gamma = 5$ and the shifting hyperparameter $\beta = 1$. The embedding size in COSFormer is set to \(d_{model} = 512\). In the Transformer Encoder \(\mathcal{E}\), we use $N_e = 2$ Transformer layers. For the Transformer Decoder \(\mathcal{D}\), we use $N_d = 2$ decoder layers. For the buffer \(\mathcal{B}_r\), the maximum number of stored WSIs is set to \(N_{buf} = 26\). To perform Eqs. \eqref{eq:text_encoding} and \eqref{eq:8}, instead of using a single class label text, we leverage the template provided by \cite{mizero}, which encodes multiple sentences for one class label. COSFormer is trained using early stopping, which is triggered when the validation loss does not decrease for five consecutive epochs. \texttt{Lookahead RAdam} \cite{lookahead} is used as the optimizer with a learning rate of $1 \times 10^{-5}$. All experiments are conducted on an NVIDIA A6000 GPU.

\subsection{Main Results} \noindent Table \ref{tab:main_results} and \ref{tab:main_results2} report the performance of COSFormer and other competing CL models across two task sequences (TCGA-CESC $\Rightarrow$ C16 and C16 $\Rightarrow$ TCGA-CESC) under both TASK-IL and CLASS-IL scenarios. 
Regardless of the CL scenarios and task sequence, COSFormer achieved the best overall performance among all competing models. For instance, in the TASK-IL scenario, COSFormer attained average accuracies of 93.137\% and 91.140\% for the two task sequences, respectively, outperforming other competing models. 
Similarly, in the more challenging CLASS-IL scenario, COSFormer was also superior to other competing models, obtaining 81.090\% and 79.359\% average accuracies across the two task sequences. These results clearly demonstrate the robustness and adaptability of COSFormer across different task orders and learning paradigms.

Although COSFormer obtained the highest average accuracy for both TASK-IL and CLASS-IL scenarios, its performance across individual tasks varied considerably. While COSFormer ranked as the best or second-best in most of the tasks, it exhibited noticeably lower performance on a few specific tasks, particularly under the CLASS-IL scenario. For example, COSFormer obtained average accuracies of 55.347\% and 47.942\% for TCGA-BRCA in the two task sequences, respectively. 
This trend of variability was also observed among all competing models, which generally exhibited even greater inconsistency in per-task performance. For instance, DER++, the second-best model in terms of average accuracy, attained average accuracies of 27.479\% and 41.136\% on TCGA-ESCA for the two task sequences under the CLASS-IL scenario. The remaining models performed even worse on both TASK-IL and CLASS-IL settings, further underscoring the difficulty of maintaining stable performance across diverse pathology tasks in continual learning environments.

In a head-to-head comparison between TASK-IL and CLASS-IL settings, the performance of COSFormer and other competing models was almost always higher in TASK-IL, highlighting the increased difficulty and complexity inherent in the CLASS-IL scenario. On average, COSFormer experienced a performance drop of 12.047\% (from 93.137\% to 81.090\%) on the C16 $\Rightarrow$ TCGA-CESC task sequence and a drop of 11.781\% (from 91.140\% to 79.359\%) on the TCGA-CESC $\Rightarrow$ C16 task sequence. It is noteworthy that other competing models generally exhibited larger drops in performance for the two task sequences: GDumb showed the largest declines of 39.740\% and 38.672\%, followed by A-GEM with drops of 32.487\% and 26.240\%, DER++ with 20.982\% and 13.195\%, and ER-ACE with 13.370\% and 11.091\%. These results further underscore COSFormer’s robustness and relative stability in the more challenging CLASS-IL setting compared to other CL methods.

\subsection{Ablation Results}

\begin{table*}[!ht]
\centering
\begin{minipage}[t]{0.48\textwidth}
\caption{Effectiveness of $\mathcal{T}$ and $WoI_\mathcal{T}$.}
\centering
\resizebox{\linewidth}{!}{\begin{tabular}{c|c|c|c}
\toprule
\textbf{Sequence} & \textbf{$\mathcal{T}$ for $EC$} & \textbf{$WoI_\mathcal{T}$ for $\mathcal{D}$} & \textbf{Avg. Acc} \\ \midrule
\multirow{4}{*}{C16 $\Rightarrow$ CESC} & \textcolor{red}{\xmark} & \textcolor{red}{\xmark} & 81.090 ($\pm$4.649) \\
 & \textcolor{teal}{\cmark} & \textcolor{red}{\xmark} & 91.931 ($\pm$2.284) \\
 & \textcolor{red}{\xmark} & \textcolor{teal}{\cmark} & 90.966 ($\pm$1.759) \\
 & \textcolor{teal}{\cmark} & \textcolor{teal}{\cmark} & \textbf{93.137 ($\pm$2.700)} \\ \midrule
\multirow{4}{*}{CESC $\Rightarrow$ C16} & \textcolor{red}{\xmark} & \textcolor{red}{\xmark} & 79.359 ($\pm$1.502) \\
 & \textcolor{teal}{\cmark} & \textcolor{red}{\xmark} & 90.770 ($\pm$1.293) \\
 & \textcolor{red}{\xmark} & \textcolor{teal}{\cmark} & 89.305 ($\pm$0.659) \\
 & \textcolor{teal}{\cmark} & \textcolor{teal}{\cmark} & \textbf{91.140 ($\pm$1.198)} \\ \bottomrule
\end{tabular}}
\label{tab:task_supplement}
\end{minipage}%
\hfill
\begin{minipage}[t]{0.48\textwidth}
\caption{Effectiveness of $EC$, $\mathcal{B}_r$, and $\mathcal{D}$.}
\centering
\resizebox{\linewidth}{!}{\begin{tabular}{c|c|c|c|c}
\toprule
$EC$ & \textbf{$\mathcal{D}$} & \textbf{$\mathcal{B}_r$} & \textbf{CLASS-IL} & \textbf{TASK-IL} \\ \midrule
\textcolor{red}{\xmark} & \textcolor{red}{\xmark} & \textcolor{red}{\xmark} & 80.950 ($\pm$1.225) & 86.066 ($\pm$0.571) \\
\textcolor{teal}{\cmark} & \textcolor{red}{\xmark} & \textcolor{red}{\xmark} & 75.825 ($\pm$2.468) & 89.070 ($\pm$3.102) \\
\textcolor{teal}{\cmark} & \textcolor{teal}{\cmark} & \textcolor{red}{\xmark} & 69.230 ($\pm$2.120) & 92.253 ($\pm$2.134) \\
\textcolor{teal}{\cmark} & \textcolor{teal}{\cmark} & \textcolor{teal}{\cmark} & \textbf{81.090 ($\pm$4.649)} & \textbf{93.137 ($\pm$2.700)} \\ \midrule
\end{tabular}}
\label{tab:ablation_results}
\end{minipage}
\end{table*}

\noindent \textbf{Investigating the Task Supplement for $\mathcal{E}$ and $\mathcal{D}$.} While the model is unaware of the target task $\mathcal{T}$ and $WoI_\mathcal{T}$ in the CLASS-IL scenario, in the TASK-IL scenario, the model is informed of $\mathcal{T}$ and $WoI_\mathcal{T}$. To investigate the effect of these two components, we conducted TASK-IL experiments with and without $\mathcal{T}$ and $WoI_\mathcal{T}$. The results are shown in Table \ref{tab:task_supplement} for both the C16 \(\Rightarrow\) CESC sequence and its reversed counterpart, CESC \(\Rightarrow\) C16. The findings indicate that the two components are essential for performance improvement, achieving its best performance, equipped with both $\mathcal{T}$ and $WoI_\mathcal{T}$, for both sequences. In a direction comparison between $\mathcal{T}$ and $WoI_\mathcal{T}$, the addition of task information $\mathcal{T}$ to $\mathcal{E}$ yielded slightly higher gains than providing $WoI$ to \(\mathcal{D}\), with improvements of \(+0.947\%\) and \(+1.465\%\) for the two sequences, respectively. 
 
\noindent \textbf{Effectiveness of $EC$, $\mathcal{D}$, and $\mathcal{B}_r$} To further assess the effectiveness of COSFormer, we evaluated the impact of three major components: 1) Expert Consultation $EC$; 2) Transformer Decoder with autoregressive decoding $\mathcal{D}$; 3) Text-based Retrieval Buffer $\mathcal{B}_r$. The results on the C16 $\Rightarrow$ CESC sequence are reported in Table \ref{tab:ablation_results}. For both scenarios, the combination of all three components (COSFormer) achieved the highest performance. However, the individual contributions of each component varied between the two scenarios.
Under the TASK-IL scenario, each component clearly demonstrated its effectiveness when added sequentially, leading to improvements of $+3.004\%$, $+6.187\%$, and $+7.071\%$ over the baseline (without these three components). In contrast, the more challenging CLASS-IL scenario revealed inconsistent effects from individual components. Specifically, adding $EC$ and $\mathcal{D}$ sequentially led to performance decreases of $-5.125\%$ and $-11.720\%$, respectively. This degradation is attributable to increased logit diversity across labels, which may introduce prediction ambiguity. Meanwhile TASK-IL remains robust due to masking of irrelevant words in $WoI_\mathcal{T}$. Importantly, COSFormer effectively mitigated such limitations observed in the CLASS-IL scenario through the Text-based Retrieval Buffer $\mathcal{B}_r$, which stores representative and diverse WSIs. By enabling the model to replay knowledge from previous tasks, it helps maintain stable performance in this challenging setting.

\begin{figure*}[!h]
\centerline{\includegraphics[width=1\textwidth]{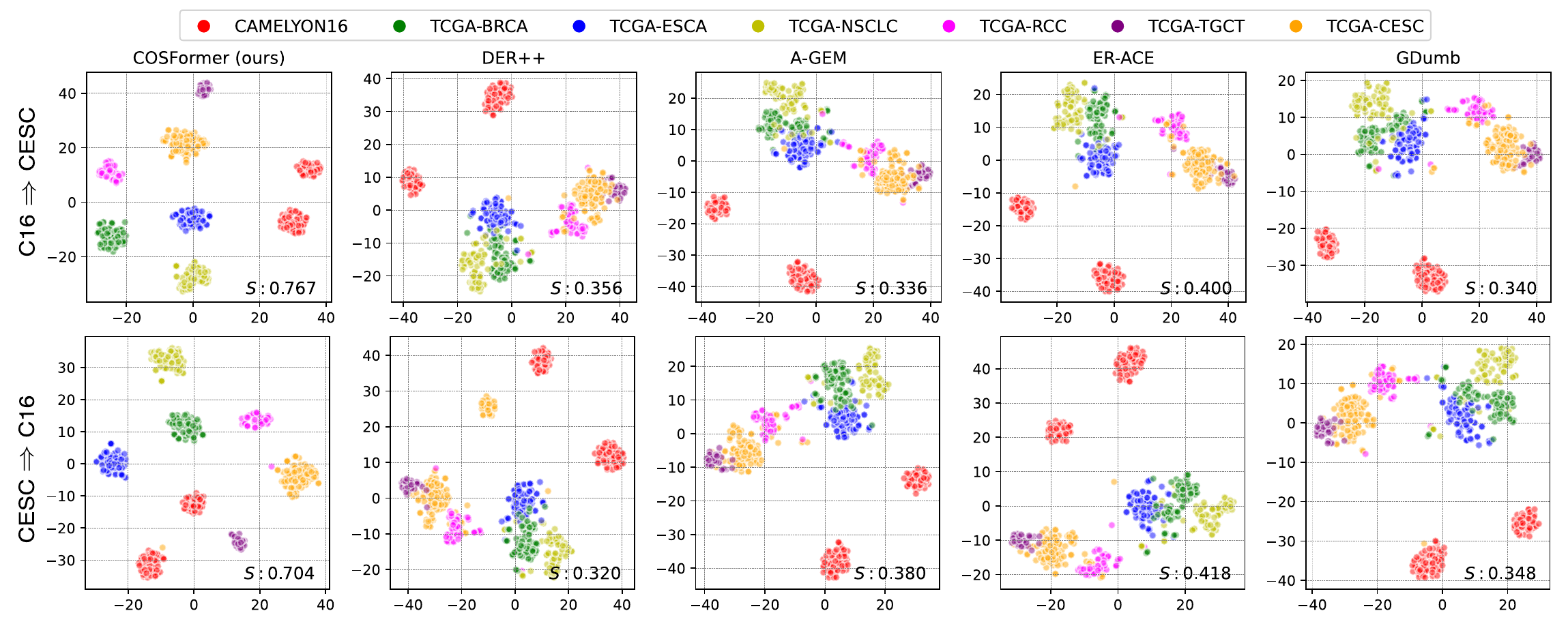}}
\caption{Comparison of WSI embeddings between COSFormer and other CL methods using t-SNE. \textit{S} denotes the Silhouette score.}
\label{fig:tsne}
\end{figure*}

\subsection{t-SNE Qualitative Results} \noindent To investigate how effectively COSFormer and other CL methods represent WSIs across tasks, we visualize their embeddings, which are used as inputs to the classification head. We adopt t-SNE \cite{tsne} to project the embeddings to a two-dimensional space, where each color represents a distinct task (Fig. \ref{fig:tsne}). It is evident that COSFormer produces embeddings that are well-separated between datasets and well-clustered within the same dataset, aligning with their respective tasks. This observation is further supported by the Silhouette score (S) \cite{silhouettes}, where COSFormer obtained the highest scores in both data sequences. This enhanced representation improves class separability, ultimately leading to superior classification performance compared to other CL methods.

\section{Discussion}

\noindent 
We propose a continuous learning mechanism tailored for diverse pathology image analysis tasks. COSFormer introduces a flexible and adaptive framework that address key limitations of conventional computational pathology approaches. In real-world pathology, the number and type of diagnostic categories can vary substantially across datasets, tasks, and clinical scenarios. By leveraging expert consultation, a text-based buffer storage, and autoregressive decoding, COSFormer facilitates efficient acquisition of new knowledge while effectively mitigating catastrophic forgetting. The superior performance of COSFormer in both TASK-IL and CLASS-IL settings suggest its potential as a practical and scalable solution for real-world pathology applications.

Following the setting established by \cite{rebuffi2017icarl}, previous studies \cite{agem,derpp,erace} commonly assume that each task contains the same number of categories and the number of output logits in the classification head is fixed at the beginning of training. However, this assumption is obviously impractical for real-world applications. Notably, fixing the number of classes restricts the adaptability of the model, as pathology tasks often involve varying numbers and types of categories. Accordingly, the number of categories and output logits should remain flexible and be defined dynamically based on the specific requirements of each task, rather than being constrained by technical limitations. COSFormer overcomes these tissues via the autoregressive decoding mechanism \cite{bui2024mecformer}. This enables the dynamic generation of diagnostic terms, allowing the model to accommodate new class labels without requiring modifications to the model architecture, thereby enhancing its adaptability and sustainability  for real-world clinical deployment. 

COSFormer effectively leverages the Mixture-of-Experts (MoE) paradigm, which assumes that each task is best handled by a dedicated expert through learnable projection weights. This design allows COSFormer to project slide embeddings into more discriminative, task-specific subspaces, thereby enhancing class separability and improving overall performance. Unlike shared feature spaces that may entangle representations from different tasks, MoE-driven feature learning helps preserve task-specific knowledge and mitigates interference between tasks.
As illustrated in Fig.~\ref{fig:tsne}, COSFormer produces embeddings that are both well-separated across tasks and tightly clustered within tasks, indicating a strong alignment with the underlying tasks. Compared to other rehearsal-based methods, which tend to produce more entangled or diffuse feature distributions, COSFormer demonstrates superior representational quality and task-awareness, further validating its effectiveness in addressing the challenges of continual learning in pathology.

The role of buffer storage and past-to-present learning strategies has proven to be critical in continual learning. Prior rehearsal-based approaches \cite{agem,derpp,erace} typically adopted a reservoir strategy, which stores a fixed number of slides and randomly replaces them with new data arrive. However, such random strategy is suboptimal for capturing the most informative and representative examples. In contrast, we employ a multimodal selection approach, based upon vision-language similarity scores, to retain slides that are best aligned with their corresponding text prompts for each task. These selected slides better capture the underlying histopathological features relevant to their tasks, leading to improved replay quality and overall performance. This is demonstrated in the ablation study reported in Tab.~\ref{tab:ablation_results}, where omitting the Text-based Retrieval Buffer Storage Strategy results in a clear drop in performance in both CLASS-IL and TASK-IL scenarios.


This study has several limitations. 
First, although COSFormer achieved the best overall performance across both TASK-IL and CLASS-IL settings, its performance on individual tasks was not always superior. In certain cases, other methods outperformed COSFormer on specific tasks, suggesting that while the model effectively balances knowledge across tasks, there may be trade-offs in optimizing performance for individual tasks. Further investigation needs to be followed to better understand the factors that influence per-task performance and to explore mechanisms that could improve performance consistency across all tasks.
Second, we evaluated COSFormer on 7 cancer subtyping datasets. While cancer subtyping is a critical problem in pathology, other tasks, such as survival prediction and treatment response prediction, also paly essential roles in clinical practice. Future work will entail the adaptation of COSFormer to a broader range of pathology tasks. 
Third, only two task sequences (C16 $\Rightarrow$ TCGA-CESC and TCGA-CESC $\Rightarrow$ C16) were investigated in this study. As the order and structure of tasks may influence continual learning performance, further work will explore COSFormer on more diverse and randomized sequences to thoroughly assess its generalizaiblity and robustness.
Fourth, as a new task is introduced, COSFOrmer requires the addition of new parameters of size \(d_f \times d_{model}\). While it is relatively small compared to the overall model size, the cumulative growth of parameters across many tasks may pose scalability concerns. To mitigate this issues, we plan to investigate more parameter-efficient approaches or dynamic parameter-sharing strategies to enhance scalability in long-term continual learning settings.
Last, COSFormer exploits the autoregressive decoding mechanism to dynamically handle various diagnostic terms. While this design enhances adaptability and flexibility, it inevitably introduces additional computational overhead. Future work will explore strategies to improve decoding efficiency and reduce computational cost.

\section{Conclusion}

\noindent In this paper, we present COSFormer, a novel framework that integrates an expert consultation, inspired by the doctor consultation process for evaluating patient samples to avoid ambiguous representations across tasks, with a Transformer-based encoder-decoder architecture with autoregressive decoding to predict class labels as diagnostic terms. We also introduce an efficient CL training framework. To the best of our knowledge, COSFormer is evaluated on the largest benchmark for CL in WSI classification, comprising a sequence of seven WSI datasets/tasks. Both qualitative and quantitative results demonstrate its superior effectiveness compared to traditional CL methods.

\section*{Acknowledgements}
This study was supported by the National Research Foundation of Korea (NRF) grant (No. 2021R1A2C2014557 and No. RS-2025-00558322).

\section*{Statements of ethical approval}
Not applicable.

\section*{Declaration of competing interest}
The authors declare that they have no conflict of interest.


\bibliographystyle{apacite}
\bibliography{bibliography.bib}

\end{document}